\documentclass[runningheads]{llncs}
\usepackage{graphicx}

\usepackage{tikz}
\usepackage{comment}
\usepackage{amsmath,amssymb} %
\usepackage{color}

\usepackage[accsupp]{axessibility}  %

\usepackage{amsmath,amsfonts,bm}

\def\eqref#1{equation~\ref{#1}}

\def\1{\bm{1}}

\newcommand{\test}{\mathcal{D_{\mathrm{test}}}}

\DeclareMathAlphabet{\mathsfit}{\encodingdefault}{\sfdefault}{m}{sl}
\SetMathAlphabet{\mathsfit}{bold}{\encodingdefault}{\sfdefault}{bx}{n}

\def\sR{{\mathbb{R}}}

\usepackage{color,xcolor}
\usepackage{epsfig}
\usepackage{graphicx}

\usepackage{adjustbox}
\usepackage{array}
\usepackage{booktabs}
\usepackage{colortbl}
\usepackage{wrapfig}
\usepackage{hhline}
\usepackage{multirow}
\usepackage{subcaption} %
\usepackage{graphics}

\usepackage{amsmath,amsfonts,amssymb}
\usepackage{bm}
\usepackage{nicefrac}
\usepackage{microtype}

\usepackage{changepage}
\usepackage{extramarks}
\usepackage{fancyhdr}
\usepackage{lastpage}
\usepackage{setspace}
\usepackage{soul}
\usepackage{xspace}
\usepackage{enumitem}

\usepackage{url}

\usepackage{algorithm, algorithmic}
\usepackage{enumitem}  %

\usepackage{makecell}

\usepackage{pifont} %

\usepackage{float}

\usepackage{tabularx}
\usepackage{longtable}

\newcolumntype{L}[1]{>{\raggedright\let\newline\\\arraybackslash\hspace{0pt}}m{#1}}
\newcolumntype{C}[1]{>{\centering\let\newline\\\arraybackslash\hspace{0pt}}m{#1}}
\newcolumntype{R}[1]{>{\raggedleft\let\newline\\\arraybackslash\hspace{0pt}}m{#1}}
\newcolumntype{Y}{>{\centering\arraybackslash}X}

\newcommand{\sect}[1]{Section~\ref{#1}}

\newcommand{\fig}[1]{Fig.~\ref{#1}}
\newcommand{\tbl}[1]{Table~\ref{#1}}

\newcommand{\degree}{\ensuremath{^\circ}\xspace}
\newcommand{\ignore}[1]{}

\DeclareMathAlphabet{\mathbfit}{OML}{cmm}{b}{it}

\makeatletter
\DeclareRobustCommand\onedot{\futurelet\@let@token\@onedot}
\def\@onedot{\ifx\@let@token.\else.\null\fi\xspace}

\def\eg{e.g\onedot} 
\def\ie{i.e\onedot}

\def\wrt{w.r.t\onedot}

\def\etal{et al\onedot}

\makeatother

\definecolor{MyDarkBlue}{rgb}{0,0.08,1}
\definecolor{MyAqua}{rgb}{0,0.7,0.7}
\definecolor{MyDarkGreen}{rgb}{0.02,0.6,0.02}
\definecolor{MyDarkRed}{rgb}{0.8,0.02,0.02}
\definecolor{MyDarkOrange}{rgb}{0.40,0.2,0.02}
\definecolor{MyPurple}{RGB}{111,0,255}
\definecolor{MyRed}{rgb}{1.0,0.0,0.0}
\definecolor{MyGold}{rgb}{0.75,0.6,0.12}
\definecolor{MyDarkgray}{rgb}{0.66, 0.66, 0.66}

\newcommand{\modelfull}{Manual-to-Executable-Plan Network\xspace}
\newcommand{\model}{MEPNet\xspace}

\newcommand{\vnew}[1]{\ensuremath{V^{\textit{new}}_{#1}}}
\newcommand{\vcur}{\ensuremath{V^{\textit{cur}}}}

\newcommand{\xhdr}[1]{{\noindent \bf #1}}

\usepackage{parskip}

\begin{document}
\pagestyle{headings}
\mainmatter
\def\ECCVSubNumber{4284}  %

\title{Translating a Visual LEGO Manual to a Machine-Executable Plan} %

\titlerunning{Translating a Visual LEGO Manual to a Machine-Executable Plan}
\author{Ruocheng Wang\inst{1}\and
Yunzhi Zhang\inst{1}\and
Jiayuan Mao\inst{2}\and \\
Chin-Yi Cheng\inst{3}\thanks{Work done when working at Autodesk AI Lab.}\and
Jiajun Wu\inst{1}
}
\authorrunning{Wang et al.}
\institute{Stanford University \and
Massachusetts Institute of Technology \and
Google Research
}
\maketitle

\begin{abstract}

\vspace{-1em}
We study the problem of translating an image-based, step-by-step assembly manual created by human designers into machine-interpretable instructions.
We formulate this problem as a sequential prediction task: at each step, our model reads the manual, locates the components to be added to the current shape, and infers their 3D poses.
This task poses the challenge of establishing a 2D-3D correspondence between the manual image and the real 3D object, and 3D pose estimation for unseen 3D objects, since a new component to be added in a step can be an object built from previous steps.
To address these two challenges, we present a novel learning-based framework, the \modelfull (\model), which reconstructs the assembly steps from a sequence of manual images. The key idea is to integrate neural 2D keypoint detection modules and 2D-3D projection algorithms for high-precision prediction and strong generalization to unseen components. The \model outperforms existing methods on three newly collected LEGO manual datasets and a Minecraft house dataset.

\end{abstract}

\section{Introduction}
\begin{figure*}[t]
    \centering
    \includegraphics[width=\textwidth]{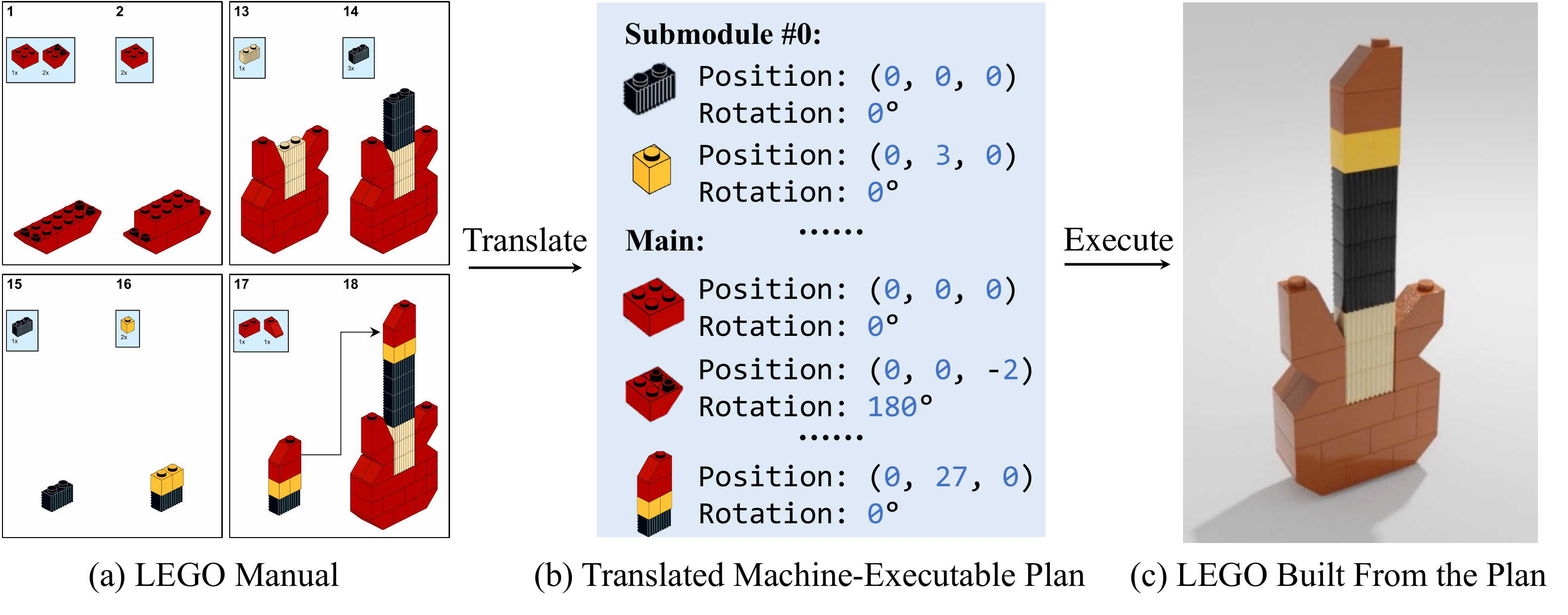}
    \vspace{-20pt}
    \caption{We study the problem of translating a LEGO manual to a machine-executable plan that can be executed to build the target shape. (a) Screenshots from an original LEGO manual. (b) A machine-executable plan generated by our model \model. (c) LEGO built by executing the generated plan.}
    \label{fig:lego-teaser}
\end{figure*}

As a community, we would like to build machines that can assist humans in constructing and assembling complex objects, such as block worlds~\cite{chen2019order}, LEGO models~\cite{chung2021brick}, and furniture~\cite{suarez2018can}. The assembly task involves a sequence of actions that move different 3D parts to desired poses. Tackling such a long-horizon task with machines requires significant engineering effort~\cite{suarez2018can}. On the other hand, humans usually rely on visual manuals to guide assembly procedures. These manuals are built by expert designers to decompose the task into a sequence of short steps that can be executed smoothly and efficiently. In this paper, we aim to facilitate the assembly tasks for machines by building a model that translates manuals into machine-interpretable plans. \fig{fig:lego-teaser}a shows an example of LEGO manuals that guides the user to build a guitar. Each step in the manual involves multiple building components presented in 2D images, and our goal is to extract the pose of each component in order to inform a downstream autonomous agent to execute the step to build the target object.

We identify two key challenges of interpreting visual manuals. First, it requires identifying the correspondence between a 2D manual image and the 3D geometric shapes of the building components. Since each manual image is the 2D projections of the desired 3D shape, understanding manuals requires machines to reason about the 3D orientations and alignments of components, possibly with the presence of occlusions. 

The second challenge is the rich library of assembly components. Taking LEGO as an example, while most LEGO shapes can be built from a finite set of primitives, these primitives can be flexibly composed into more complex subparts that are added to the main body as a whole (\eg, the head of the guitar in \fig{fig:lego-teaser}) in a step. The compositionality of primitive bricks greatly increases the diversity of LEGO components, and as a result increases the difficulty for machines to interpret LEGO manuals: it requires inferring 3D poses of {\it unseen} objects composed of seen primitives.

In this work, we develop a method that tackles this challenging problem. Concretely, we formulate the problem of translating manuals into machine-executable plans as a sequential task. For each step, the inputs consist of 1) a set of primitive bricks and parts that have been built in previous steps represented in 3D, and 2) a target 2D image showing how components should be connected to each other. The expected output is the (relative) poses of all components involved in this step, as shown in \fig{fig:lego-teaser}b.

There are roughly two groups of solutions to parsing a single step of a manual. The first group is search-based methods (\ie,  ``analysis-by-synthesis'')~\cite{yuille2006vision,bever2010analysis}. These methods make use of a given forward synthesis model, \ie, the underlying manual image renderer, for pose inference. They search over possible 3D poses of new components, render manual images based on the candidate poses, and select the pose that maximizes the matching score between the input and the rendered images.
This approach is simple and accurate but assumes a given renderer. It is also computationally expensive as we need to search for the 3D poses of multiple components jointly in a single assembly step. The second group is learning-based, using end-to-end neural networks that predict the 3D pose of each component. This approach does not require an image renderer and is fast, but typically suffers from poor generalization to unseen 3D shapes. 

Inspired by these observations, we propose the \modelfull (\model), a hybrid approach that combines the best of both worlds. The \model has two stages. In the first stage, a convolutional neural network takes as input the current 3D LEGO shape, the 3D model of new components, and the 2D manual image of the target shape. It predicts a set of 2D keypoints and masks for each new component. In the second stage, 2D keypoints predicted in the first stage are back-projected to 3D by finding possible connections between the base shape and the new components. It also refines component orientation predictions by a local search. Our approach does not require the groundtruth image renderer during training or inference. Experiments show that our proposed approach maintains the efficiency of learning-based models, and generalizes better to unseen 3D components compared to end-to-end learning-based approaches.

We evaluate \model on two benchmarks: one in the LEGO domain and another in a Minecraft-style house crafting domain~\cite{Chu2016HouseCraft}. Our results show that \model enables more accurate pose estimation and, more importantly, generalizes better to unseen novel 3D components compared with several baselines. Most notably, we demonstrate that \model is capable of generalizing to real-world LEGO manuals by training solely on synthetically generated manuals. We will release all code and data for full reproducibility.

\section{Related Work}
\xhdr{Parsing human-designed diagrams.} A diagram is a fundamental and commonly-used tool for humans to communicate concepts and information \cite{agrawala2011design}. There have been a number of works on parsing different types of diagrams like engineering drawings~\cite{haralick1982understanding}, cartographic road maps~\cite{mena2003state} and sewing patterns~\cite{berthouzoz2013parsing} into machine-understandable data. We focus on the task of parsing assembly manuals into instructions that can be executed by machines.  Shao~\etal~\cite{shao2016dynamic} proposes a technique to parse assembly instructions into executable plans, but focuses manuals in vector-graphic formats, while we work on LEGO manuals in RGB image format with no known visual primitives like edges and polygons. Li~\etal~\cite{li2020learning} proposes a method to parse the 3D poses of parts from a single RGB image, while our work focuses on sequentially inferring 3D poses from a series of manual images.

\xhdr{Inverse 3D modeling.} Inferring the geometry procedures that reconstruct a 3D shape is useful in many domains like robotic assembly~\cite{suarez2018can,lee2021ikea}, shape synthesis~\cite{funkhouser2004modeling,chaudhuri2011probabilistic,li2017grass} and computer-aided design~\cite{willis2021fusion,li2020sketch2cad}. A line of research focuses on using different geometry operations like poses of 3D primitive parts~\cite{abstractionTulsiani17}, shape programs~\cite{tian2018learning,jones2020shapeassembly}, constructive solid geometry~\cite{du2018inversecsg} and CAD operations~\cite{xu2021inferring}. Different kinds of information are studied to guide the inverse inference process: final 3D shape~\cite{mo2019structurenet,huang2020generative}, single image~\cite{niu2018im2struct,li2020learning} or multi-view images~\cite{van2015part,chung2021brick}. When it comes to image-guided inverse modeling, previous works often assume access to images of the final 3D shape, where the inverse problem is inherently ambiguous. To tackle this issue, a shape prior is learned from a corpse of 3D shapes like ShapeNet~\cite{chang2015shapenet}, which can not be directly transferred to distribution different from the training data. Our work focuses on recovering the poses of LEGO bricks from a series of manual images that progressively specify building operations, which are intended to guide the assembly of diverse shapes of objects. 

\xhdr{Pose estimation.} To parse the information from manual images, we need to infer the poses of primitive parts to be assembled. Estimating the 3D poses of objects is a fundamental problem in 3D vision. A line of research uses convolutional neural networks to directly localize objects in a scene and regress their 6D poses~\cite{brachmann2014learning,xiang2017posecnn,li2018deepim,wang2019densefusion}. Other works adopt a two-stage approach where 2D keypoints of objects are first extracted and then poses are inferred from them~\cite{rad2017bb8,oberweger2018making,peng2019pvnet}. Most works focus on detecting objects from the same category of the training objects, while our work aims to build models that can estimate the pose of known primitives as well as novel shapes composed from them. Xiao~\etal~\cite{xiao2019pose,xiao2020few} proposes a CNN architecture to estimate the pose of a single object in the image with a known 3D model. On the other hand, assembly manuals often require estimating poses of multiple objects. Furthermore, to understand manuals, models need to ignore the parts that are assembled in previous steps and only detect parts of interest, which is not addressed in previous works.

\section{Problem Formulation}
Throughout the paper, we will be using LEGO manuals as our example, although many of the ideas generalize to other types of assembly manuals such as Minecraft and furniture. A LEGO manual is composed of a sequence of images. The images specify a step-by-step instruction sequence of adding new components to an existing LEGO shape (called the {\it base shape}, which is typically the target shape in the previous step). Each new component is either a primitive brick with specified type and quantities, (in which case the image will specify the type and the number of new bricks) or a pre-built component (called a {\it submodule}, such as the guitar head shown in \fig{fig:lego-teaser}). The main diagram in each image will specify the poses of the new components \wrt the {\it base shape}, by showing a 2D view of the target shape at this step.

\xhdr{Sequential manual parsing.} We formulate the task as a sequential prediction task of $T$ steps. At each step, a manual parser receives the following inputs.
\vspace{0.5em}
\begin{enumerate}[noitemsep,topsep=0pt,parsep=0pt,partopsep=0pt,leftmargin=2em]
    \item A 3D representation of the {\it base shape} at this step. In this paper, we focus on a voxel-based representation $V^\textit{cur}_i$. It might be an empty voxel grid when we start building a new shape. In LEGO manuals, this usually comes from the target shape of the previous step.
    \item A set of components to be added at this step, where each component is either a primitive brick or a pre-built shape composed of multiple primitive bricks. In either cases, their shape will be represented as a set of 3D voxels $V^{\textit{new}}_{ij}$, where $j=1,2,\cdots,C_i$ and $C_i$ is the number of components for step $i$.
    In a LEGO manual image, these are usually specified in a small diagram at a corner as shown in \fig{fig:lego-teaser}a.
    \item A 2D image $I_i$ specifying the target shape after we compose all components at step $i$. In LEGO manuals, this corresponds to the main diagram of a manual image.
\end{enumerate}
\vspace{0.5em}
Our goal is to predict the 3D translation and rotation \wrt $V_i^\text{cur}$ for each added component $\{(t_{ij}, r_{ij})\}_{j=1}^{C_i}$ for each step $i$, where $t_{ij}\in\sR^{3}, r_{ij}\in\sR^{3\times3}$. 

\begin{figure}[!t]
    \centering
    \includegraphics[width=\textwidth]{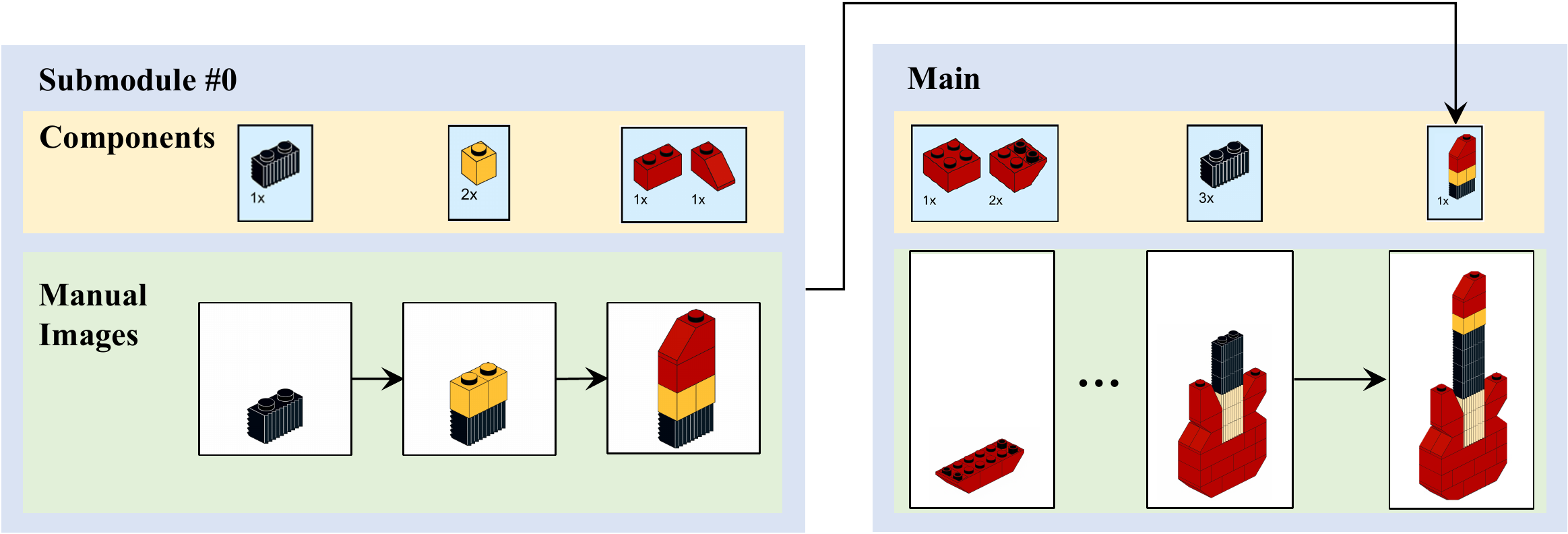}
    \captionof{figure}{Our factorized representation of LEGO manuals, which is a tree-structured plan specified by manual images. Each component is either a primitive brick or a submodule, and each submodule is recursively constructed from a sequence of manual images with corresponding components. }
    \vspace{-1em}
    \label{fig:factorized}
    \vspace{-1em}
\end{figure}

Once we have a model that predicts 3D poses of components at each step, they serve as a plan that can be executed by machines to assemble a complete object (called a LEGO {\it set}) in an iterative manner. In general, the dependency between steps is a tree because of {\it submodules} in assembly---a step may depend on the resulting shapes from multiple previous steps, as shown in \fig{fig:factorized}.

\xhdr{The connection constraint.} An important feature of object assembly is the inherent connection constraints among object parts, such as the studs in LEGO shapes and the pegs and holes in furniture assembly. This enables designers to simply use a 2D image to specify how parts should be connected, in contrast to specifying the exact 3D dimensions and poses of objects.
This constraint brings us benefits in both model design and evaluation. For example, in model design, it is generally hard to accurately infer the 3D continuous pose of an object from a 2D image, but it is easier to infer a set of discrete connections between objects. On the evaluation side, this allows us to reconstruct a physically plausible (stable and no inter-penetration) target shape by simulating all steps.

\begin{figure*}[t]
    \centering
    \includegraphics[width=\textwidth]{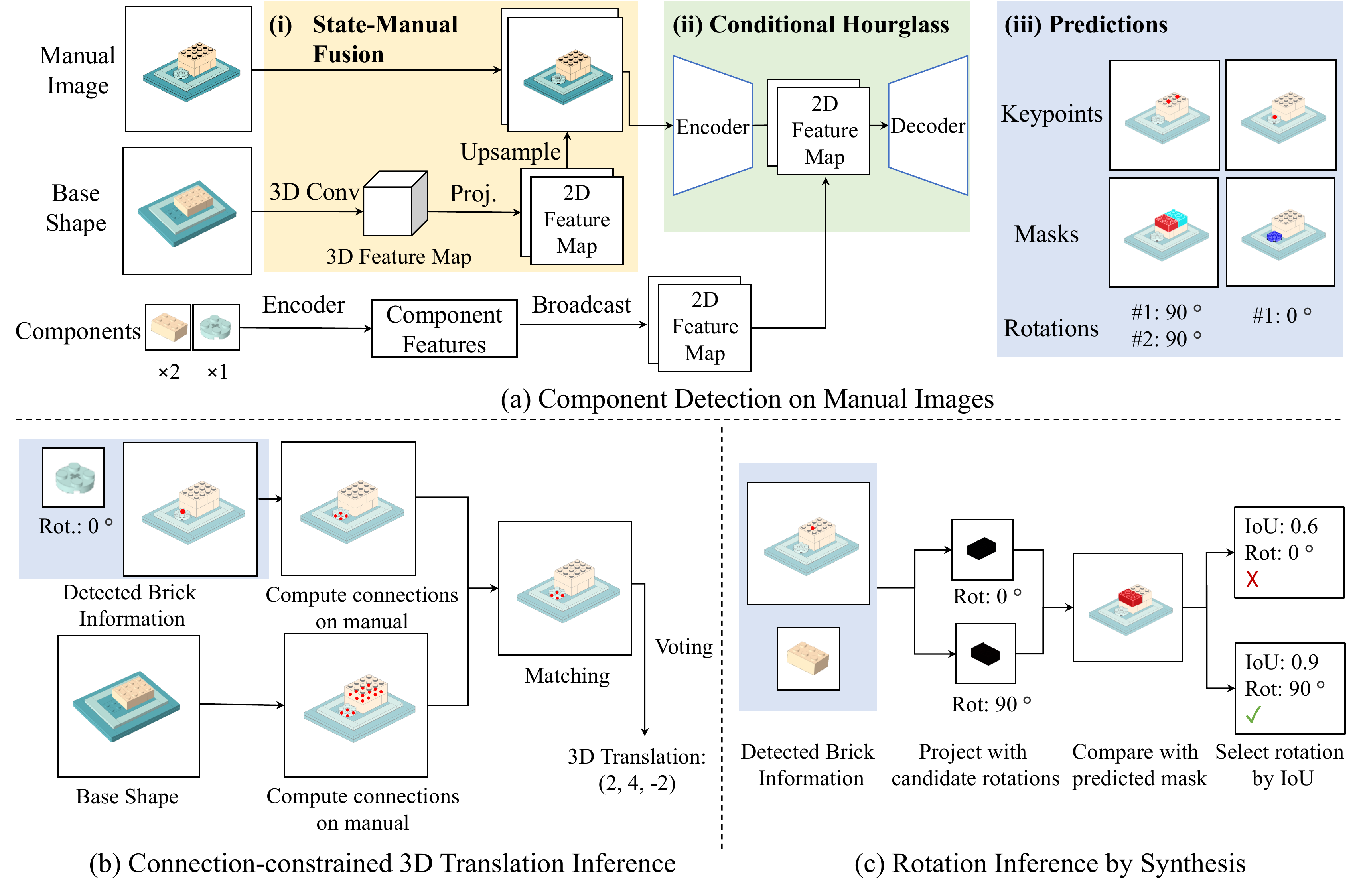}
    \vspace{-20pt}
    \caption{\model consists of two stages. (a) In the first stage, the model detects the 2D center keypoints, the masks, and the rotations for all added components on the manual. (b) and (c): We use a connection-constrained inference subroutine and an inference-by-synthesis subroutine to recover 3D poses of components from 2D information predicted in the first stage.}
    \vspace{-1.5em}
    \label{fig:model_fig}
\end{figure*}
\vspace{-5pt}
\section{\modelfull}
\vspace{-5pt}
In this section, we will be focusing on developing a learning-based method for solving the one-step prediction task. It can be applied iteratively to each assembly step to reconstruct the full 3D shape. Recall that the input to our model is the base shape at this step $V^{\textit{cur}}$, a set of new components, $\{ V^{new}_j \}$, and a 2D image $I$ (omitting the step index $i$ for all variables for clarity). Our goal is to infer the 3D poses of all components $\{ V^{new}_j \}$.

Our model, the \modelfull (\model), consists of two stages. In the first stage, we use a neural pose estimation model to predict the 2D keypoint (corresponding to the center of the object), mask, and rotation for each new component to be added. In the second stage, we use two deterministic algorithms to infer the 3D poses of new components by fusing the predictions from the first stage.

Our two-stage approach is primarily motivated by the difficulty of directly estimating 3D poses from 2D images, especially considering the generalization to unseen components. Here, we leverage the connection constraint between components and the idea of analysis-by-synthesis to design post-processing algorithms that refine the raw prediction made by neural networks. In the experiment section, we will compare our two-stage approach with other alternatives, including a single-stage neural network baseline and an analysis-by-synthesis baseline based on mask predictions.

\vspace{-1em}
\subsection{Neural Pose Estimation}
\label{sec:neuralpose}
Our neural module for estimating component poses follows an encoder-fusion-decoder pipeline. It combines convolution-based 3D shape encoders and a 2D Hourglass network---a state-of-the-art model for 2D keypoint and mask prediction. We will use separate 2D and 3D encoders to extract features for \vcur, \vnew{j}, and $I$, fuse the latent representations, and predict the center keypoint, mask, and rotation for each individual $\vnew{j}$.

Our first module, shown in \fig{fig:model_fig}a (i) is a state-manual encoder, which fuses the information of \vcur and the image $I$ representing the target shape. Specifically, the state-manual encoder takes the voxel representation of $V^\textit{cur}$ as input, and extracts a 3D representation with two 3D Convolution-BatchNorm-ReLU modules. Let $f_{3d}$ denote the output tensor of shape $\sR^{C_1 \times H\times W\times D}$, where $(H, W, D)$ is the 3D dimension of the input voxel and $C_1$ is the number of channels. Then, we use the camera parameters of the manual image to transform this voxel to the camera frame.
This is computed by a differentiable spatial transformation based on \cite{jaderberg2015spatial}.
Next, we project this voxel to the camera plane by rasterization, that is, for each pixel on the sensor plane, select the first non-empty voxel hit by a ray shooting from the sensor pixel towards the camera center. This gives us a 2D feature map $f^{2d} \in \sR^{C_1\times H\times W}$. Then the feature map is upsampled with bilinear interpolation to have the same resolution as the manual image $I$. Then we concatenate this feature map and the image along the feature channel to generate an ``augmented'' image representation, \ie, $I^{a} \in \sR^{(C_1+3) \times H_\text{img} \times W_\text{img}}$.

Our second module, shown in \fig{fig:model_fig}a (ii) is a component-conditioned Hourglass model for predicting the center keypoint, mask, and rotation for each component. Specifically, for each new component \vnew{j}, also represented as a 3D voxel, we use five 3D Convolution-BatchNorm-ReLU layers, followed by an average pooling layer to extract the corresponding feature, denoted as $f^{\textit{new}}_j \in \sR^{C_2}$. Note that, for multiple components that have the same shape (\wrt SO(3)), we only encode one of them. We order all components based on their order in the input manual and get an sequence of component feature embeddings $\{f^{\textit{new}}_j\}$. We concatenate all embeddings along the channel dimension into a vector $f^{\textit{new}*}$, whose number of channels is $K \times C_2$, where $K$ is the total number of distinct-shaped components, which we call ``{\it component types}''. We pad this vector by adding $0$'s into a vector of length $K_\textit{max} \times C_2$, where $K_\textit{max} = 5$ is the maximum number of distinct components considered in \model.

Given the state-aware manual image $I^a$ and component features $\{f^\textit{new}_j\}$, we predict the 3D poses of the added components using an adapted implementation of stacked Hourglass Networks~\cite{newell2016stacked,zhou2019objects}. We first use a top-down network to process the $I^a$ into a low-resolution 2D feature map $I^\textit{aenc}$ of resolution $\left(H_\text{img}/32,W_\text{img}/32\right)$. Next, we tile the concatenated component embeddings $f^{\textit{new}*}$ along spatial dimensions into a 2D feature map of resolution $\left(H_\text{img}/32,W_\text{img}/32\right)$, and concatenate this feature map to $I^\textit{aenc}$. Then we use a bottom-up decoder network to output a high-resolution feature map of resolution $\left(H_\text{img}/4,W_\text{img}/4\right)$. %

Then we use three separate small fully-convolutional neural networks to extract the center keypoint, mask, and rotation for each input component based on the feature map output by the Hourglass decoder.
\vspace{0.5em}
\begin{enumerate}[noitemsep,topsep=0pt,parsep=0pt,partopsep=0pt,leftmargin=2em]
    \item For the center keypoint, we employ the structure of CenterNet~\cite{zhou2019objects}. For each component type, the model output is a tuple of three 1D feature maps: $(h, \textit{dx}, \textit{dy})$. $h$ is a heatmap of centers, and $\textit{dx}$ and $\textit{dy}$ form a two-dimensional offset prediction which is the difference between the actual center of an object and the pixel location of the heatmap. This helps mitigate the discretization error caused by downsampling.
    \item For the component mask, we employ the structure of Associative Embedding~\cite{newell2017associative}. For each component type, the output is a tuple of two feature maps $(m, \textit{emb})$, where $m$ is a segmentation mask of the component type and $\textit{emb}$ is a 2D feature map where each pixel is associated with a vector embedding (called the ``associative embeddings''). The L2 distance between pixels associated with different instances should be large, while the distance between pixels of the same instance should be small. To get the instance-level segmentation, we perform a pixel clustering based on $\textit{emb}$.
    \item  For the rotation, our model output is a 4-dimensional vector, with a Softmax nonlinearity\footnote{In the implementation we used a slightly more complex scheme to handle symmetries. Details are included in the supplementary material.}, since the component will only have rotations chosen from (0\degree, 90\degree, 180\degree, 270\degree) around the vertical axis in the LEGO problem we considered.
\end{enumerate}

It is important to note that since we have merged components of the same shape into one component type, there will be multiple instances detected for each component type. For example, shown in \fig{fig:model_fig}a, the center keypoint heatmap for the first component type has two peaks, corresponding to two instances of this component type.

\xhdr{Training and losses.} We train \model with full supervision on a synthetically generated dataset where we have the groundtruth keypoint, mask, and rotation information. The entire neural network module is trained end-to-end with gradient descent. Our objective function is computed by
\begin{center}
\vspace{-2.0em}
\begin{equation*}
    \mathcal{L} = \alpha\cdot \mathcal{L}_\text{keypoint} +\beta \cdot \mathcal{L}_\text{mask} + \gamma\cdot\mathcal{L}_\text{rotation}.
\end{equation*}
\vspace{-2.0em}
\end{center}
We adopt the keypoint loss adapted from \cite{zhou2019objects}: $\mathcal{L}_\text{keypoint} = \mathcal{L}_\text{heatmap} + \mathcal{L}_\text{offset}$, where $\mathcal{L}_\text{heatmap}$ is a focal loss~\cite{lin2017focal} computed based on the predicted heatmap and a groundtruth heatmap generated by Gaussian kernels. $\mathcal{L}_\text{offset}$ is an L1 loss for the regression task of $\textit{dx}$ and $\textit{dy}$ (the offsets). The mask loss is adapted from \cite{newell2017associative}: $\mathcal{L}_\text{mask} = \mathcal{L}_\text{semantic} + \mathcal{L}_\text{pull} + \mathcal{L}_\text{push}$, where $\mathcal{L}_\text{semantic}$ is a cross-entropy loss applied to the predicted mask, and $\mathcal{L}_\text{pull} + \mathcal{L}_\text{push}$ is the contrastive loss for learning the associative embeddings. Finally, we use a cross-entropy loss to train the rotation prediction module.  More details are in the supplementary material. 

\subsection{3D Pose Inference}
\label{sec:3d_infer}
Based on the 2D predictions from the first stage, we infer 3D poses for each component. Here, we will exploit two important ideas: the connection constraint in assembly domains, and the idea of analysis-by-synthesis.

\xhdr{Connection-constrained 3D translation inference.} Given the center keypoint of each component, our goal is to find the 3D XYZ position of the component. Here, we will rely on the connection constraints, that is, there should be at least one position where the new component is attached to another existing brick. In LEGO, the attachment is achieved by inserting a ``stud'' of the existing brick into an ``anti-stud'' of the new component. As illustrated in \fig{fig:model_fig}b, our inference has three steps. First, given the center keypoint of the new component, we infer the 2D location of all anti-studs of the new component (this process is deterministic and does not require any depth information because images in LEGO manuals are orthographically projected, see a detailed proof in the supplementary material). Next, we also project all studs in the base shape $\vcur$ onto the same 2D plane. Finally, we perform a matching between all possible studs and anti-studs followed by a majority voting. Then we can predict the 3D position of the new component based on the 3D position of existing studs, which is known.

\xhdr{Rotation inference by synthesis.} 
We empirically found that directly predicting the rotation of a {\it submodule} is hard. Thus, instead of using the rotation predicted from \sect{sec:neuralpose}, we employ an analysis-by-synthesis process to estimate the rotation. Illustrated in \fig{fig:model_fig}c, for each {\it submodule}, 
we compute all possible translations for each of the 4 rotations based on connection constraints. Then for each rotation-translation candidate $(r, t)$, we project the component with pose $(r, t)$ to the image, obtain a mask and compute its Intersection-over-Union (IoU) with the predicted mask from \sect{sec:neuralpose}. We select the pose $(r, t)$ with the highest IoU score as the final pose prediction.

\subsection{Implementation Details}
\xhdr{LEGO discretization.} To enable flexible attachments between different LEGO bricks, most LEGO bricks have sizes that are the multiples of the smallest $1\times1\times1$ brick and thus are inherently voxelized. In this work, we voxelize the basic $1\times1\times1$ LEGO brick as a $2\times2\times2$ voxel grid. We double the resolution of the voxel grid because the 3D translation of a brick can be half the size of the brick.

\xhdr{Center keypoint.} We observe that although the components of interest can be severely occluded in a manual image, their top faces often remain visible. This is a consistent design pattern across many LEGO manuals~\cite{heiser2004identification}. Thus, the center keypoint of a primitive brick is defined as the center of the top surface of a LEGO brick, as illustrated in \fig{fig:model_fig}a (iii). For submodules, we define the center keypoint to be the center keypoint of the topmost primitive brick in the submodule. If there are multiple topmost primitive bricks, we use a randomly selected brick during training and use a modified 3D pose inference algorithm during evaluation. Details are in supplementary materials.

\xhdr{Camera projection.} Same as the real-world LEGO manuals, we model the camera projection as a weak perspective (scaled orthographic) transformation. Thus, the camera is parameterized by scale $s\in \sR$, translation $t\in\sR^2$, and rotation $r$ represented by three Euler angles. In this paper, we assume known camera parameters for all methods including baselines. The camera parameters are predicted by a pretrained pose estimation model \cite{xiao2019pose} which takes the input base shape and manual image as input. Details are in supplementary materials.

\section{Experiments}
\label{sec:exp}
We evaluate \model in two assembly domains: LEGO and 3D-Craft~\cite{chen2019order}.

\subsection{Setup}

\xhdr{Baselines.} We compare \model with two baselines.

PartAssembly~\cite{li2020learning} is a two-stage method designed for inferring object part poses from a single image. It first predicts the masks for individual components. Then, it encodes the feature for each component and predicts the pose. We have made the following adaptations to it for our sequential prediction setting. First, to incorporate information about the current object state, we replaced the input image with the state-augmented manual image $I^a$ using the same encoder as \model. Second, we add CoordConv~\cite{liu2018intriguing} to the model to enhance its prediction about 3D positions. Third, we add a post-processing procedure to the model prediction that quantizes the prediction based on the connection constraints. Finally, we also replaced their original point cloud encoder with our voxel-based encoder. More details can be found in the supplementary materials. 

Direct3D, the second baseline, is an ablative variant of \model, in which we directly predict the 3D translation and rotation for each instance (instead of predicting keypoints, masks, and rotations). We also use CoordConv and prediction quantization with connection constraints in this model.

\xhdr{Metrics.}
We evaluate \model and baselines at three different levels of granularity: componentwise, stepwise and setwise. In the componentwise and the stepwise case, the input to each model is the ground truth voxel grid $V^{cur}$ and submodules (if any). We evaluate the {\it 3D pose accuracy} (correct or incorrect because we have quantized the model predictions using the connection constraint) and the {\it Chamfer distance}~\cite{fan2017point}. To account for the rotation symmetry of components, we restricted the set of possible rotations based on component shapes. For stepwise pose accuracy, we say a step is correct if the predictions for all components in this step are correct.
To compute Chamfer distance, following~\cite{wang2018pixel2mesh}, we uniformly sample 10000 points from their meshes.
For stepwise Chamfer distance, we compute the metric between the union of all components in a step. 

In the setwise case, each model will be run on all manual images sequentially, and auto-regressively. That is, the predicted target shape from the first step will be the input for the second step. Submodules will be built by models as well. We compute two metrics: the Chamfer distance between the ground truth final shape and the shape output by each model. We also compute a normalized {\it Mistakes to Complete (MTC)} score, proposed in \cite{chen2019order}. MTC computes the average percentage of steps where a model gives wrong poses predictions. In this case, the model will be fed with the ground truth $V^{cur}$ and submodules.

\begin{figure}[t]
    \centering
    \includegraphics[width=\linewidth]{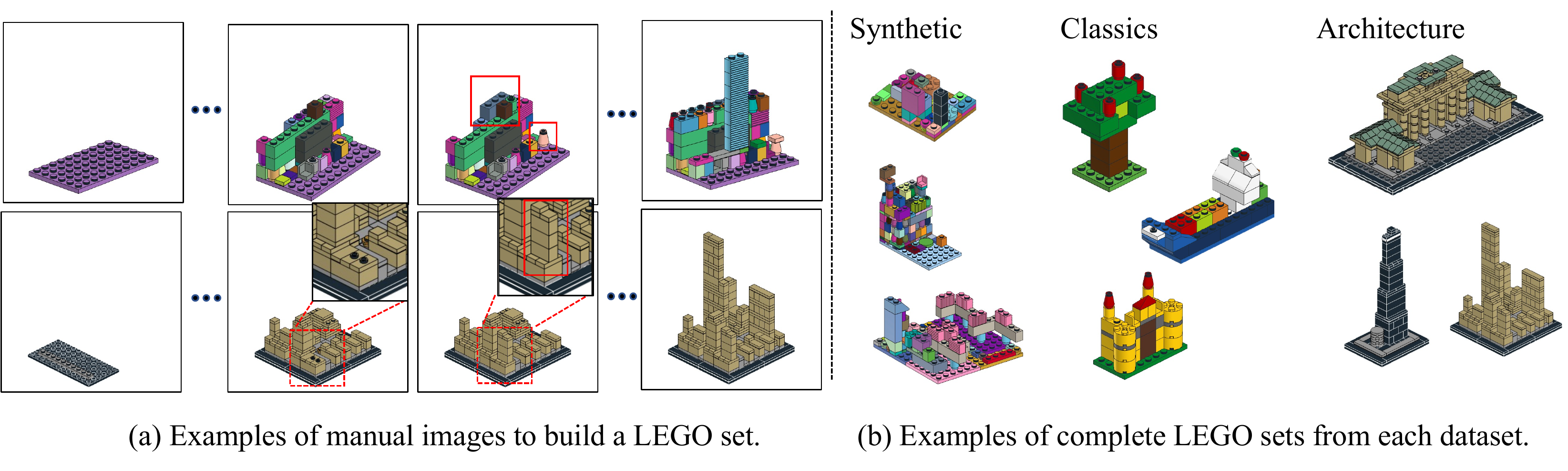}
    \vspace{-2em}
    \caption{Example manual images and shapes in our LEGO datasets.}
    \label{fig:dataset_ex}
\end{figure}

\subsection{Results on LEGO}

\xhdr{Dataset.}
Our first dataset is a synthetic LEGO dataset that is procedurally generated based on 72 types of primitive bricks, and rendered using a standard LEGO manual renderer\footnote{\url{https://trevorsandy.github.io/lpub3d/}}. Our data generation pipeline encapsulates two features of real-world LEGO manuals: 1) using submodules to assemble two components that have been built separately, and 2) stacking or tiling multiple bricks of the same shape at one step to form structures such as walls and floors. Examples of generated LEGO objects are shown in \fig{fig:dataset_ex}(b). 
Details of the generation pipeline, attribution of assets and their license can be found in the supplementary materials. 
We generate 8000 manuals for training, 10 sets for validation, and 20 sets for testing. In sum, there are 200K individual steps in training, 300 for the validation split, and 600 for the test split.

We have also collected two datasets from real-world LEGO manuals. We select 11 sets from LEGO's Classics theme, which contains simple objects designed for children older than 4, and 5 sets from LEGO's Architecture theme, which contains more complex building-shaped LEGOs for kids older than 10. There are around 200 individual steps in each dataset. We manually exclude or replace bricks that are not in our primitive brick set, and re-render the LEGO manuals with the factorized representation.
Examples of our datasets are shown in \fig{fig:dataset_ex}.

\begin{figure*}[t]
    \centering
    \includegraphics[width=\linewidth]{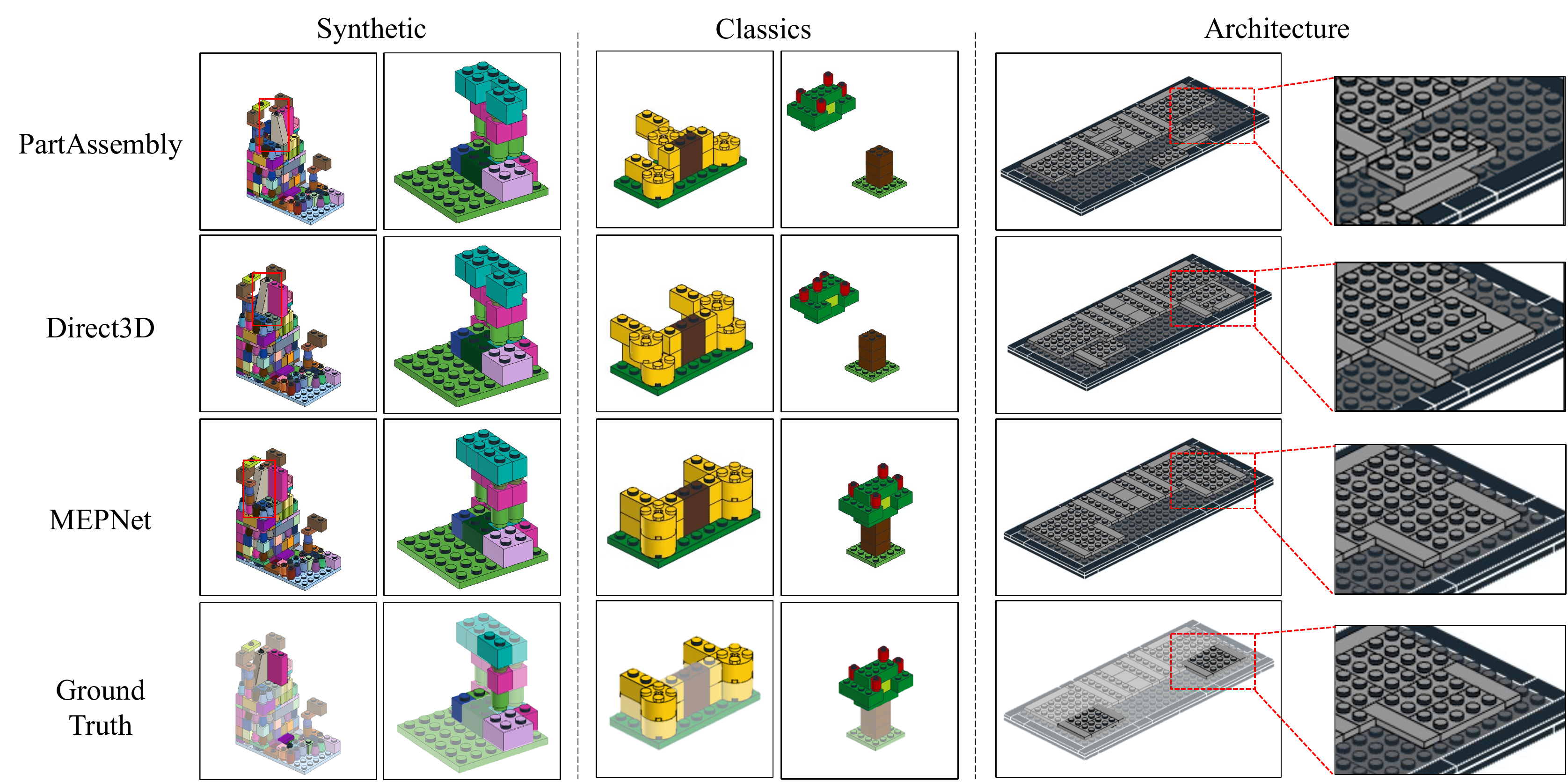}
    \vspace{-15pt}
    \caption{Qualitative Results on the LEGO datasets. Each column contains the ground truth and the predictions from models for a single step. Components added in the step are highlighted in the manual images. To have a straightforward comparison between different models, we render their predictions in the same way as rendering the target manual. }
    \vspace{-.5em}
    \label{fig:step_vis}
\end{figure*}
\begin{table}[!t]
\centering
\scriptsize
\begin{tabularx}{\textwidth}{llcccccc}
\toprule
\multicolumn{1}{l}{}          & \multicolumn{1}{l}{} & \multicolumn{2}{c}{Componentwise} & \multicolumn{2}{c}{Stepwise}           & \multicolumn{2}{c}{Setwise} \\
\cmidrule(lr){3-4}\cmidrule(lr){5-6}\cmidrule(lr){7-8}
\multicolumn{1}{l}{}          & \multicolumn{1}{l}{} & Pose Acc $\uparrow$ (\%)   & CD $\downarrow$     & Pose Acc $\uparrow$ (\%)& \multicolumn{1}{l}{CD $\downarrow$} & CD $\downarrow$  & MTC $\downarrow$ (\%)    \\
\midrule
\multirow{3}{*}{Synthetic}    & PartAssembly~\cite{li2020learning}         & $63.74$           & $393.21$     & $47.05$        & $226.48$                 & $321.91$            & $52.72$  \\
                              & Direct3D             & $88.51$           & $32.60$      & $77.39$        & $5.08$                   & $9.18$              & $25.62$  \\
                              & \model                 & $\bf 96.96$           & $\bf 11.60$      & $\bf 93.41$        & $\bf 1.93$                   & $\bf 4.74$              & $\bf 8.63$   \\
\midrule
\multirow{3}{*}{Classics}     & PartAssembly~\cite{li2020learning}         & $2.26$            & $1171.57$    & $0.00$         & $296.82$                 & $775.29$            & $100.00$ \\
                              & Direct3D             & $34.84$           & $303.56$     & $24.27$        & $3.00$                   & $67.75$             & $80.77$  \\
                              & \model                 & $\bf 88.69$           & $\bf 72.79$      & $\bf 90.29$       & $\bf 0.10$                   & $\bf 15.52$             & $\bf 13.97$  \\
\midrule
\multirow{3}{*}{Architecture} & PartAssembly~\cite{li2020learning}         & $5.12$            & $1964.31$    & $3.24$         & $858.58$                 & $719.88$            & $96.13$  \\
                              & Direct3D             & $13.71$           & $936.77$     & $23.17$        & $227.41$                 & $191.28$            & $87.79$  \\
                              & \model                 & $\bf 83.47$           & $\bf 136.40$     & $\bf 82.23$        & $\bf 15.35$                  & $\bf 107.95$            & $\bf 16.00$ \\
\bottomrule
\end{tabularx}
\vspace{5pt}
\caption{Quantitative results of models on the three LEGO datasets. Chamfer distance metrics are multiplied by a factor of $10^5$. \model outperforms baselines in all metrics on the three datasets.}
\vspace{-2.5em}
\label{tbl:res_main}
\end{table}

\begin{figure*}[t]
    \centering
    \includegraphics[width=\linewidth]{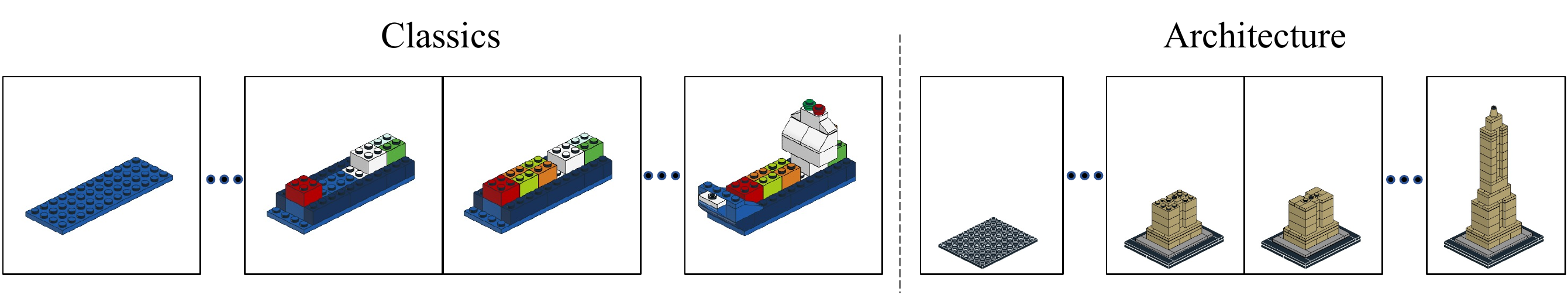}
    \vspace{-20pt}
    \caption{Qualitative Results of \model building LEGOs from scratch. We visualize several intermediate steps and final results by rendering them in the same way as the manual images. \model can successfully parse manuals into executable plans with diverse target shapes.}
    \label{fig:setwise_vis}
\end{figure*}

\xhdr{Results on the synthetic dataset.}
Quantitative results on the synthetic dataset are summarized in \tbl{tbl:res_main}. \model outperforms baseline models in all metrics considered. From visualizations in \fig{fig:step_vis}, we can see \model is able to accurately predict 3D poses of both primitive bricks and submodules, even in cases with significant occlusions. Baseline models tend to either have a small deviation in 3D translation, or fail to infer the orientation of submodules. We also find that PartAssembly fails to predict accurate masks for target components, which are critical in its pose estimation.

\xhdr{Generalization to the Classics and the Architecture datasets.}
We directly apply models trained on our synthetically generated datasets on the Classics and the Architecture datasets. Illustrated in \tbl{tbl:res_main}, there is a large performance drop for PartAssembly and Direct3D due to the distribution mismatch between these two datasets and our synthetic training dataset, while \model maintains high accuracy. We found that the Direct3D model can still predict accurate keypoints on these two datasets, but fails to accurately predict their 3D poses. This shows the effectiveness of our two-stage inference procedure. Noticeably, \model can successfully build sets with diverse shapes guided by manuals as illustrated in \fig{fig:setwise_vis}, although they look drastically different from the synthetic dataset and contain submodules of unseen shapes. We present several examples of the full assembly process in the supplementary materials.

The architecture dataset is the most challenging one because it contains a large number of primitive bricks and severe occlusion. \model still achieves a decent prediction accuracy while both baselines fail significantly.

\begin{table}[!t]
\setlength{\tabcolsep}{3pt}
\centering
\scriptsize
\begin{tabularx}{\textwidth}{lcccccc}
\toprule
\multicolumn{1}{l}{} & \multicolumn{2}{c}{Synthetic} & \multicolumn{2}{c}{Classics}           & \multicolumn{2}{c}{Architecture} \\
\cmidrule(lr){2-3}\cmidrule(lr){4-5}\cmidrule(lr){6-7}
\multicolumn{1}{l}{} & Pose Acc $\uparrow$ (\%)   & CD $\downarrow$     & Pose Acc $\uparrow$ (\%)   & CD $\downarrow$  & Pose Acc $\uparrow$ (\%)   & CD $\downarrow$ \\
\midrule
PartAssembly~\cite{li2020learning}  & $0.00$    & $1754.64$   & $0.00$   & $2107.11$     & $0.00$    & $4816.21$     \\
Direct3D               & $22.75$   & $161.59$ & $10.00$   & $180.86$  & $0.00$    & $1539.04$ \\
\model (w.o. RS)       & $22.27$          & $192.93$ & $20.00$          & $25.17$ & $23.08$          & $\bf 256.55$ \\
\model              & $\bf 75.83$   & $\bf 23.27$ & $\bf 90.00$   & $\bf 0.25$ & $\bf 38.46$   & $271.35$ \\
\bottomrule
\end{tabularx}
\vspace{5pt}
\caption{Quantitative results on the submodules exclusively. Chamfer distance metrics are multiplied by a factor of $10^5$. Rotation inference by synthesis (RS) plays an important role in inferring the pose of submodules.}
\label{tbl:res_submodules}
\vspace{-1em}
\end{table}

\xhdr{Ablation: submodules.} We perform an ablation study comparing how different models handle submodules, as shown in \tbl{tbl:res_submodules}. Specifically, we evaluate the componentwise pose prediction accuracy and Chamfer distance across all steps that involve submodules.
The table indicates the challenge of 3D pose inference for submodules. Our rotation-inference-by-synthesis (RS) consistently improves the results across all three datasets.

\begin{table}[!t]
\setlength{\tabcolsep}{2.2pt}
\centering
\scriptsize
\begin{tabularx}{\textwidth}{lcccccc}
\toprule
\multicolumn{1}{l}{} &  \multicolumn{3}{c}{Classics}           & \multicolumn{3}{c}{Architecture} \\
\cmidrule(lr){2-4}\cmidrule(lr){5-7}
\multicolumn{1}{l}{}   & Pose Acc $\uparrow$ (\%) & Time $\downarrow$ (s)&  TLE $\downarrow$ (\%)  & Pose Acc $\uparrow$ (\%) & Time $\downarrow$ (s)&  TLE $\downarrow$ (\%)\\
\midrule
Direct3D              & $34.84$   & ${\bf 0.233}$ & ${\bf 0}$   & $13.71$  & ${\bf 0.320}$    & ${\bf 0}$ \\
\model (a-by-s)       & $48.42$          & $10.94$ & $0.9$          & $34.14$ & $70.77$          & $11.65$ \\
\model              & {\bf 88.69}   & {0.311} & {\bf 0}   & \bf 83.47   & {0.406} & {\bf 0} \\
\bottomrule
\end{tabularx}
\vspace{5pt}
\caption{Comparison with an ablation model based purely on analysis-by-synthesis. We show the component pose accuracy and the average inference time per step (which may include multiple components). TLE (Time-Limit-Exceeded) measures the percentage of components whose prediction cannot terminate in 1 minute.}
\vspace{-2em}
\label{tbl:res_aofs}
\end{table}
\xhdr{Ablation: pure analysis-by-synthesis.} We also perform an ablation study of the second stage algorithm by replacing it with an algorithm that is entirely based on analysis-by-synthesis, for both primitive bricks and submodules. In contrast, our full model \model only applies analysis-by-synthesis for the rotation inference of submodules. Since we do not assume access to the underlying image renderer, we will perform analysis-by-synthesis based on the 2D mask of the manual image.

Specifically, we use the mask prediction from our Hourglass model. Our goal is to set the 3D poses for new components to maximize the matching score between detected 2D masks and the 2D projections of these new components. To avoid the exponential scaling with respect to the number of components, we employed a sequential greedy algorithm. Given the current shape and a set of new components to be added, we iteratively search the pose for each new component. For each component, we enumerate all possible poses, and select the pose that minimizes the IoU between the projected mask and segmentation mask of that component type predicted by the Hourglass model. We set the maximum searching time for each component to 1 minute.

The accuracy and runtime of different models are summarized in \tbl{tbl:res_aofs}. Direct3D is the fastest because 3D poses are directly predicted by the Hourglass model, but the performance is significantly worse than other methods. The analysis-by-synthesis baseline outperforms the Direct3D method, but it runs significantly slower. Our full model, \model, performs the best. We empirically attribute the inferior performance of the analysis-by-synthesis variant to the imprecise prediction of component masks, especially for small primitive bricks. 

\subsection{Results on 3D-Craft}
\begin{figure}[!t]
    \centering
    \includegraphics[width=\linewidth]{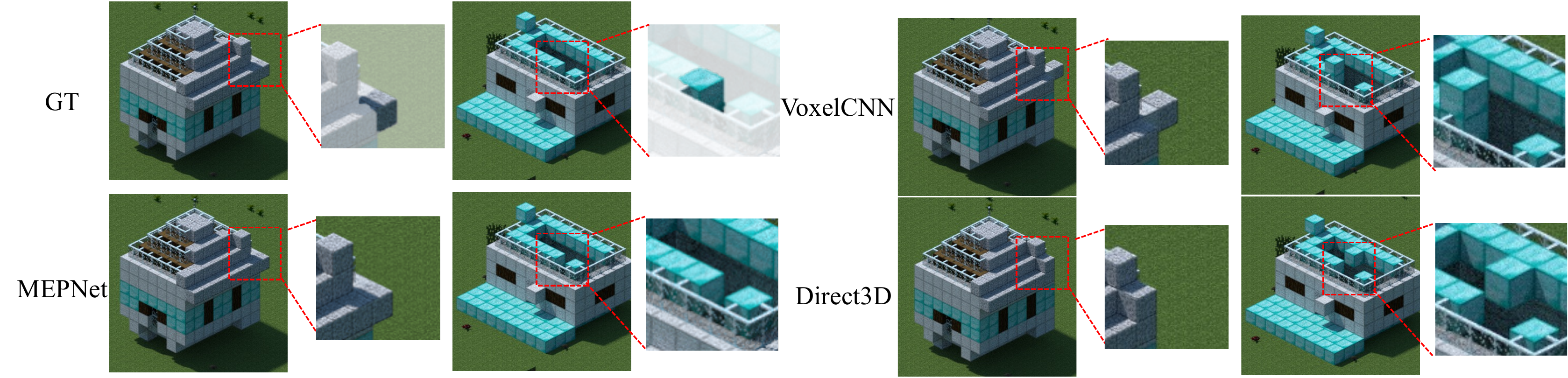}
    \caption{Qualitative results on the 3D-Craft dataset. Ground truth and predictions of new bricks are zoomed in. Predictions of baseline models often lie in a small neighborhood of the ground truth positions, while our model is more accurate.}
    \vspace{-1em}
    \label{fig:craft_vis}
\end{figure}
We also evaluate \model on building Minecraft houses from 3D-Craft~\cite{chen2019order}. 

\xhdr{Data.}
The original 3D-Craft contains houses built with equally-sized bricks from crowdworkers in the format of building operation sequence. We leverage this sequence information to generate manuals. Each step in the manuals contains only one new brick. And our goal is to predict the 3D translation of the new brick in each manual image. We select 80 houses (20k steps) for training, 5 (1.5k steps) for validation, and 5 (1.5k steps) for testing. As the original dataset has a heavily imbalanced distribution of brick types, we select 5 frequently-used types in the dataset. In general houses in 3D-Craft contains more bricks than LEGO, and the appearance of bricks will be affected by lighting.

\xhdr{Setup.} To migrate \model to this setting, we use one-hot embedding to encode different types of bricks occupying each voxel. Because there is no rotation involved, we also remove the rotation prediction modules for all methods.
Finally, we modify the 2D-to-3D algorithm according to Minecraft's connection constraints. Details are in the supplementary material.

\xhdr{Results.} \model achieves translation accuracy of $86.3\%$ on the 3D-Craft dataset, while VoxelCNN and Direct3D only achieve $49.8\%$ and $75.8\%$.  Based on the visualizations shown in \fig{fig:craft_vis}, we can also see that \model can yield more accurate results across all different lighting conditions.

\section{Discussion}
We study the problem of translating an image-based, step-by-step assembly manual created by human designers into machine-interpretable instructions. We propose \modelfull (\model), a model that reconstructs the assembly steps from a sequence of manual images. The key idea behind our model is to combine learning-based methods and inference-by-synthesis algorithms to wire in the connection constraints in assembly domains. Results show that our model outperforms existing methods on three newly collected LEGO manual datasets and a Minecraft house dataset. 

\vspace{2pt}
\noindent {\bf Acknowldegements:} We thank Joy Hsu, Chengshu Li, and Samuel Clarke for detailed feedback on the paper. This work is partly supported by Autodesk, the Stanford Institute for Human Centered AI (HAI), the Stanford Center for Integrated Facility Engineering (CIFE), ARMY MURI grant W911NF-15-1-0479, NSF CCRI \#2120095, the Samsung Global Research Outreach (GRO) Program, and Amazon, Analog, Bosch, IBM, Meta, and Salesforce.

\bibliographystyle{splncs04}
\bibliography{jiajun,reference}
\end{document}


\pagestyle{headings}
\mainmatter
\def\ECCVSubNumber{4284}  %

\title{Supplementary Materials: Translating Visual LEGO Manuals to a Machine-Executable Plan} %

\author{Ruocheng Wang\inst{1}\and
Yunzhi Zhang\inst{1}\and
Jiayuan Mao\inst{2}\and \\
Chin-Yi Cheng\inst{3}\thanks{Work done when working at Autodesk AI Lab.}\and
Jiajun Wu\inst{1}
}
\authorrunning{Wang et al.}
\institute{Stanford University \and
Massachusetts Institute of Technology \and
Google Research 
}
\titlerunning{Supplementary Materials}
\maketitle

First, we introduce our data generation pipeline for LEGO synthetic dataset in \sect{sec:synthetic_gen} and 3D-Craft in \sect{sec:craft_gen}. Then we provide more implementation details and hyperparameters of models in \sect{sec:impl}. \sect{sec:3d_infer_details} explains some details of the 3D Pose Inference stage of \model. \sect{sec:camera} explains some details of camera parameter estimation of our datasets. \sect{sec:add} presents some additional ablation studies. \sect{sec:limits} discusses the limitations of our paper. 
\sect{sec:attr} lists the attribution and license for assets used in this paper. Finally, we provide some step-by-step building visualization of our model in \sect{sec:vis}.

\section{Synthetic Dataset Generation}
\label{sec:synthetic_gen}

We generate the synthetic dataset in two stages: the first stage (the ``forward'' stage) randomly builds a complete LEGO shape, and the second stage (the ``backward'' stage) decomposes the final shape into multiple synthetic manual assembly steps. 

\subsubsection{Forward Stage}
In the forward stage, we first sample the length, width and maximum height of the bounding box of the final shape. Next, we repeat the following process iteratively.

At each iteration, we randomly partition the bounding box into smaller boxes, and apply one of the following operations to each box:
\begin{enumerate}
    \item Randomly fill the box with as many arbitrary components as possible. 
    \item Sample edges from the box, ``grow'' the height of the edges by adding random components. Afterwards, fill the non-edge area within the region with random components.
\end{enumerate}
To add components, each time we randomly sample a position first, and then use one of the three strategies: 1. randomly select a primitive brick and add it at the sampled position ; 2. build a submodule by randomly stacking multiple bricks within a small 3D bounding box and add it at the sampled position; 3. randomly select a primitive brick and stack multiple instances together and add them at the sampled position. We use rejection sampling to ensure each component will be connected to at least one of the other components and there is no collision between components.

\subsubsection{Backward Stage}

The backward stage constructs the manual iteratively. Starting from the fully built shape, at each iteration, we identify all the components that are removable from the current {\it base shape}, and compute the percentage of pixels of the components that are visible from a top-down view and a side view. 
If the visibility percentage is larger than a threshold, we mark the component as visible. We group the components that are removable and visible into multiple chunks, such that each chunk has at most 10 components instances and at most 5 component types. Each chunk will correspond to a step in the manual. Then we remove one chunk at a time and render the resulting shape as a manual image using predefined camera parameters. We perform this operation until all the components have been removed and obtain a series of manual images that illustrate the whole assembly process of the given object.

\section{3D-Craft Dataset Generation}
\label{sec:craft_gen}
Different from the data generation pipeline of the synthetic LEGO dataset, where we randomly select components to be in a step in the backward stage, the houses in the 3D-Craft dataset already come with a construction order made by humans. To maintain this order information, we start from an empty world, and add only one brick at a time according to the human construction sequence. To make sure that the manual image contains information about the added brick in each step, we iterate over a predefined set of viewpoints and select the viewpoint where the brick is visible. If there is no valid viewpoint we just randomly select one from a predefined set. Next, we render the scene with the selected viewpoint and use the result as the manual for this step.

\section{Implementation Details}
\label{sec:impl}
\subsection{Symmetry-aware rotation prediction.} In the LEGO bricks we considered, every brick may have rotation symmetry of order $n = 1$, $2$, or $4$. That is, the brick remains unchanged when being rotated for $\frac{360}{n}$ degrees. For each brick, our model predicts the component's rotation as well as the rotation symmetry class it belongs to. This is achieved by a 7-way classification where the prediction $c$ is interpreted as:
\begin{itemize}
    \item $c=0$: The brick has rotation symmetry of order 4. 
    \item $c=1$: The brick has rotation symmetry of order 2 and its rotation is $0\degree$.
    \item $c=2$: The brick has rotation symmetry of order 2 and its rotation is $90\degree$.
    \item $c>2$: The brick has no rotation symmetry and its rotation is $(c - 3) \cdot 90\degree$.
\end{itemize}

\subsubsection{Baseline models on LEGO datasets.} As baseline models directly predict 3D translations for each component as a continuous parameter, we implement a post-processing algorithm to refine the predictions by utilizing connection constraints in LEGOs. First, we round each entry of the predicted 3D translation to the nearest $0.5$. Then we search the translation that satisfies connection constraints in a small neighborhood of the rounded 3D translation. More specifically, the component must be connected to another component and there is no collision between components. Suppose the rounded translation is $(x, y, z)$, the neighborhood is defined as 
$$\{(x + \delta_x, y + \delta_y, z + \delta_z) |\quad\delta_i \in \{0, -0.5, 0.5, -1, 1\} \}$$
We select the valid translation from the neighborhood that is closest to the original rounded prediction. If none of the translations is valid, we use $(x, y, z)$ as the final prediction.

\subsubsection{3D-Craft dataset.} In the 3D-Craft dataset, a brick can either be put on the ground or attached to a neighboring brick. Therefore, given the base shape of the house, we can compute all 3D valid positions for the new brick. We project them onto the image plane, and perform a matching with the detected keypoint of the new brick. The keypoint of a brick is set to be its center/origin. The predicted 3D translation is set to be the valid position whose 2D projection is closest to the detected keypoint. For baseline models, we use a similar post-processing algorithm as the algorithm for LEGO datasets based on a local search.

\subsubsection{Hyperparameters} The size of the voxel grid for the base shape is set to be $130\times 130\times 130$, and the size of the voxel grid for each component is set to be $65\times 65\times 65$. For the training losses:
\begin{center}
\vspace{-1.0em}
\begin{equation*}
    \mathcal{L} = \alpha\cdot \mathcal{L}_\text{keypoint} +\beta \cdot \mathcal{L}_\text{mask} + \gamma\cdot\mathcal{L}_\text{rotation}.
\end{equation*}
\vspace{-1.0em}
\end{center}
We set $\alpha$ and $\beta$ to $1$ and $\gamma$ to $0.1$. For Direct3D, $L_\text{keypoint}$ and $\mathcal{L}_\text{mask}$ are replaced with $L_\text{trans}$ whose weight is also set to 1. All models except VoxelCNN are trained with the AdamW~\cite{loshchilov2017decoupled} optimizer. VoxelCNN is trained with SGD, following the implementation of the original paper. Other hyperparameters can be found in \tbl{tbl:hyperparams}.
\begin{table}[]
\centering
\footnotesize
\begin{tabularx}{0.7\linewidth}{ccccc}
\toprule
Dataset                    &              & Learning Rate & Batch Size & Epochs \\
\midrule
\multirow{3}{*}{LEGO}      & \model       & $0.00025$       & $16$        & $15$              \\
                           & Direct3D     & $0.00025$       & $16$         & $15$              \\
                           & PartAssembly & $0.001$        & $8$         & $15 $             \\
                           
\midrule
\multirow{3}{*}{3D-Craft} & \model       & $0.00025$      & $8$          & $10$             \\
                           & Direct3D     & $0.00025$       & $8$          & $10$              \\
                           & VoxelCNN     & $0.01$          & $8$          & $10$             \\
\bottomrule
\end{tabularx}
\caption{Hyperparameters of models.}
\label{tbl:hyperparams}
\end{table}
\section{Details of 3D Pose Inference.}
\label{sec:3d_infer_details}
\subsubsection{Inferring 2D locations of anti-studs of the new component.} Here, we show that under scaled orthographic projection,  the 2D offset on the image plane between a component's keypoint and its anti-studs can be computed without knowing the actual ``depth'' (which is unknown) of the component. This enable us to compute the 2D positions of anti-studs on the manual image only based on 2D keypoint information.

Recall that in scaled orthographic projection, the 2D projection of a point $v$ can be computed by 

\[
v' = A K_{int} K_{ext}v = \begin{pmatrix}
1 & 0 & 0\\
0 & 1 & 0\\
\end{pmatrix}
\begin{pmatrix}
s & 0 & 0 \\
0 & s & 0 \\
0 & 0 & 1 \\
\end{pmatrix}
\begin{pmatrix}
R &T \\
\end{pmatrix}v
\]
where $s\in\sR$ is the scale factor and $R\in\sR^{3\times3}, T\in\sR^{3}$ correspond to the extrinsic parameters of the camera (rotation and translation, respectively). Denote $P = A K_{int} K_{ext}$. For any two points $v_0, v_1 \in \sR^4$ in the world homogeneous coordinates, their offset on 2D image plane coordinate system can be computed as $Pv_0 - Pv_1 = P(v_0 - v_1)$.

Therefore, if we consider two points on a component (\eg, the detected keypoint and an anti-stud of the brick), their relative position on the 2D image plane can be computed independent of the placement of this component. Thus, given a detected keypoint on the 2D image plane, we can directly computes the position of all anti-studs of the component (on 2D), and match them with the studs on the base shape.

\subsubsection{Empty base shape} At the beginning of assembly, rather than connecting to another component, a component can also be put directly on ground. For this special case, given the 2D keypoint and rotation of a detected component, we will directly search over all possible translations $(x, 0, y)$ and select the one whose keypoint position is closest to the detected keypoint.

\subsubsection{Handling submodules with multiple topmost primitive bricks.} To predict the 3D translation of a submodule, the pose estimation module is expected to detect its 2D keypoint, which is the 2D keypoint of the topmost primitive brick in this submodule. However, the identity of the topmost primitive can be ambiguous when there are multiple components at the same ``top'' layer. To handle this, we extend the rotation inference by synthesis algorithm. For each submodule, we perform a joint search of the 3D rotation and the topmost brick that our model detects. Assuming there are $n$ topmost bricks for the target submodule, this results in at most $4n$ unique candidate 3D poses. Next, following the basic rotation inference by synthesis algorithm, we use a camera projection subroutine to compute the mask of the component with each candidate pose, and select the pose that has the highest IoU with the predicted mask.

\section{Camera Parameters}
\label{sec:camera}
\subsubsection{Camera parameters of the LEGO datasets.} The camera parameters in training and test sets are chosen from real LEGO manuals. Specifically, we have selected a small set of LEGO manuals, together with their corresponding target 3D shapes. Next, we use SoftRasterizer~\cite{liu2019soft} to find the camera parameters that maximize the IoU between the projected 3D shapes and the shape masks segmented from the real LEGO manuals.
We select a common set of camera parameters to render the Classics and Architecture datasets.
For the synthetic dataset, the camera parameters have scales sampled from $[1, 5]$, translations sampled from $[-1, 1]$ in the NDC space, and rotations sampled from $(0, 225^\circ\pm 10^\circ, 30^\circ \pm 10^\circ)$. This covers the camera parameter distribution of the real LEGO manuals we consider.

\subsubsection{Camera parameter estimation.} Our model can be integrated with external camera parameter estimation algorithms. Specifically, we use the model from Xiao~\etal~\cite{xiao2019pose} that receives a 3D shape and target image as inputs and predicts their camera parameters. Since the original model only predicts orientation of the camera given a target image and 3D object, we add scale and translation prediction modules using multi-layer perceptrons. The model is supervised by a mean squared error loss. We train the model using camera parameters from the synthetic datasets. Using the predicted camera parameters, \model can achieve the same pose accuracy and Chamfer distance metrics as using the groundtruth camera parameters acorss all three LEGO datasets.

\section{Additional Results}
\label{sec:add}
\begin{table}[t]
    \begin{minipage}[t]{.45\linewidth}
        \scriptsize
          \centering
      \begin{tabular}{lcc}
        \toprule
       & Pose Acc $\uparrow$ (\%)& CD $\downarrow$ \\
        \midrule
        MEPNet &        $\textbf{88.69}$  &  $\textbf{72.79}$   \\
        ResNet+DeConv\cite{Xiao_2018_ECCV} & $78.28$ & $96.43$\\
        \bottomrule
    \end{tabular}
      \caption{Comparison with an ablation model where we replaced the Hourglass network with a encoder-decoder network using ResNet and deconvolutional layers~\cite{Xiao_2018_ECCV}. Chamfer distance metrics are multiplied by a factor of $10^5$.}
        \label{tbl:supp_resnet}
    \end{minipage}%
    \quad\quad
    \begin{minipage}[t]{.45\linewidth}
        \scriptsize
      \centering
      \begin{tabular}{lcc}
        \toprule
       & Pose Acc $\uparrow$ (\%) & CD $\downarrow$  \\
        \midrule
        MEPNet (GT Trans.) & $\textbf{96.38}$      &  $\textbf{32.92}$ \\
        MEPNet (GT Rot.)  & $90.58$     &      $67.29$     \\
        \bottomrule
        \end{tabular}
        \caption{An ablation study where we use ground-truth translation or rotation to replace \model's predictions. It's harder for \model to predict translation than rotation. Chamfer distance metrics are multiplied by a factor of $10^5$.}
        \label{tbl:supp_gt}
    \end{minipage} 
\end{table}
\subsubsection{Ablation of the Hourglass architecture.} To validate the effectiveness of the Hourglass architecture, we use another encoder-decoder model built from ResNet and deconvolutional layers \cite{Xiao_2018_ECCV} to build an ablation model. We replace the Hourglass architecture in \model with this ablation model and evaluate it on the Classics dataset in terms of component-wise pose accuracy and Chamfer Distance. Results are summarized in \tbl{tbl:supp_resnet}, which show that Hourglass architecture outperforms the ResNet-based architecture.
\subsubsection{Pose accuracy with groundtruth translations or rotations.} To better understand how translation and rotation predictions contribute to the prediction error, we evaluate the component-wise pose accuracy of \model on the Classics dataset by replacing translation or rotation prediction with ground truth. Results are shown in \tbl{tbl:supp_gt}. In general, it is harder for \model to infer component translations than rotations.

\section{Discussions and Limitations}
\label{sec:limits}
A limitation of our work is that our proposed method relies on the specific connection constraints in the LEGO domain. Still, we bring this information into our pipeline motivated by the fact that connection constraints are ubiquitous in assembly domains, such as those of furniture and electrical devices. Accurate estimation of 3D poses from 2D images is generally hard. Thus, building models that leverage connection constraints are preferred. Intuitively, these constraints reduce the number of possible relative poses. Furthermore, our analysis-by-synthesis approach for leveraging constraints to post-process 3D poses is generic, although the exact detail for handling different types of connection constraints (\eg, peg-and-hole, mortise-and-tenon) may vary across domains. How to encode and leverage a richer sets of connection constraints in 3D~\cite{jones2021automate,willis2021joinable}, is a potential future work. Second, current settings only consider discrete domains. It remains to be explored how to extend our model to domains with continuous action space such as furniture assembly. Finally, in our current formulation, the prediction error will accumulate after each step. Leveraging multi-step information for joint prediction is also an important future direction.

\section{Attribution and License for Assets}
\label{sec:attr}
The 3D data files of LEGO bricks are obtained using ImportLDraw \footnote{\url{https://github.com/TobyLobster/ImportLDraw}} which extracts brick information from LDraw \footnote{\url{https://ldraw.org/parts/latest-parts.html}}. Some of the step information of real-world LEGO sets are obtained from LDraw Model Repository\footnote{\url{https://omr.ldraw.org/}}. Manuals images are rendered using LPub3D \footnote{\url{https://trevorsandy.github.io/lpub3d/}} and Bricklink Studio \footnote{\url{https://www.bricklink.com/v3/studio/download.page}}. All data
and software used in our project has been ethically collected from online resources with Creative Commons
or other open licensing terms. 
\section{Step-by-Step Building Visualization}
\label{sec:vis}
\begin{figure*}[t]
    \centering
    \includegraphics[width=\linewidth]{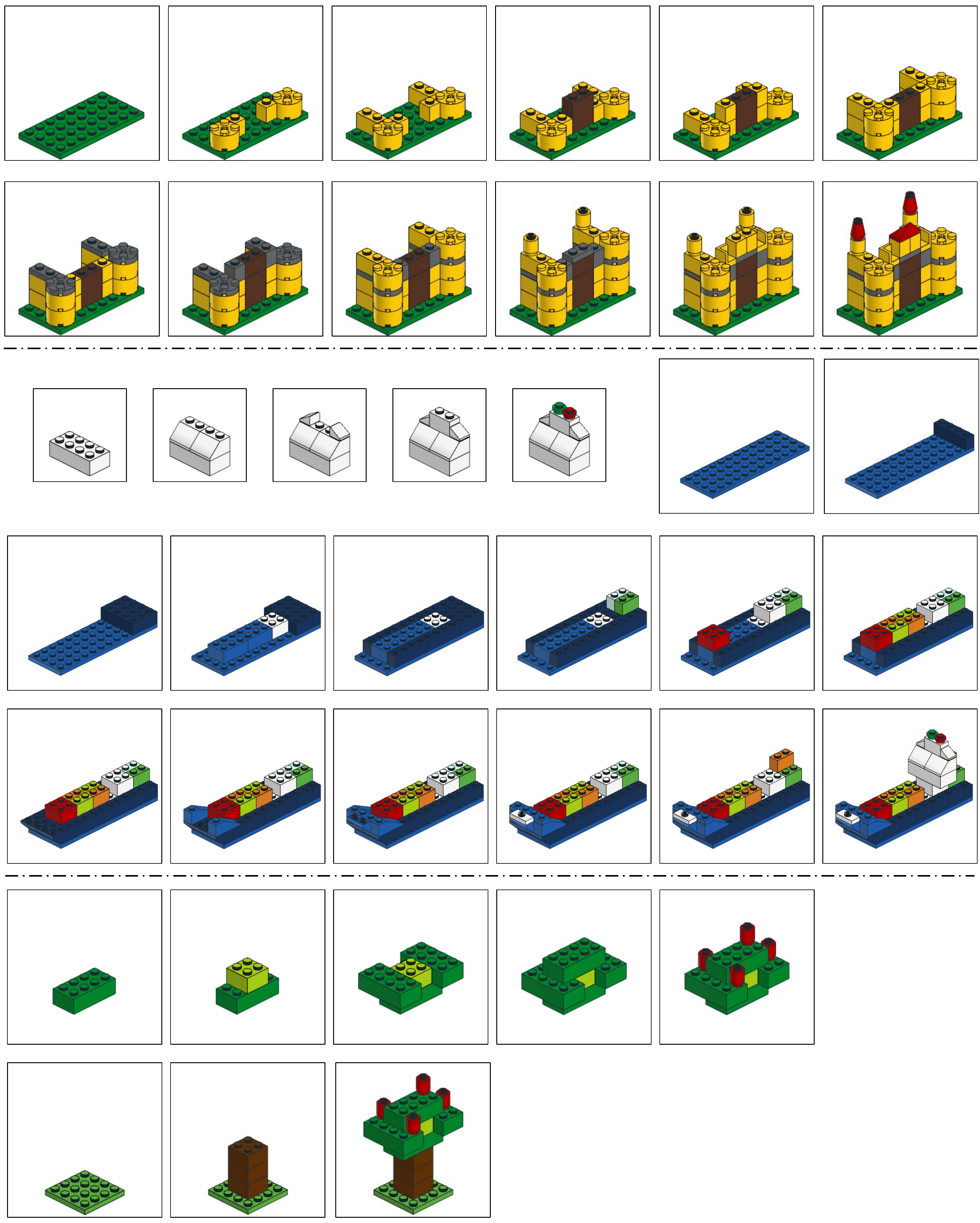}
    \vspace{-12pt}
    \caption{Visualization of the full assembly process by \model. Assembly of submodules are shown at the beginning of each set. }
    \label{fig:full_vis}
\end{figure*}

We visualize some sets that our model achieves 100\% accuracy in \fig{fig:full_vis}.

\bibliographystyle{splncs04}
\bibliography{jiajun,reference}